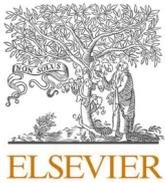
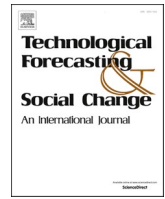

# Explainable Artificial Intelligence (XAI) from a user perspective: A synthesis of prior literature and problematizing avenues for future research

AKM Bahalul Haque [a,*], A.K.M. Najmul Islam [a], Patrick Mikalef [b,c]

[a] *Department of Software Engineering, LUT University, Finland*
[b] *Department of Computer Science, Norwegian University of Science and Technology, Norway*
[c] *Department of Technology Management, SINTEF Digital, Norway*

A B S T R A C T

The rapid growth and use of artificial intelligence (AI)-based systems have raised concerns regarding explainability. Recent studies have discussed the emerging demand for explainable AI (XAI); however, a systematic review of explainable artificial intelligence from an end user's perspective can provide a comprehensive understanding of the current situation and help close the research gap. The purpose of this study was to perform a systematic literature review of explainable AI from the end user's perspective and to synthesize the findings. To be precise, the objectives were to 1) identify the dimensions of end users' explanation needs; 2) investigate the effect of explanation on end user's perceptions, and 3) identify the research gaps and propose future research agendas for XAI, particularly from end users' perspectives based on current knowledge. The final search query for the Systematic Literature Review (SLR) was conducted on July 2022. Initially, we extracted 1707 journal and conference articles from the Scopus and Web of Science databases. Inclusion and exclusion criteria were then applied, and 58 articles were selected for the SLR. The findings show four dimensions that shape the AI explanation, which are format (explanation representation format), completeness (explanation should contain all required information, including the supplementary information), accuracy (information regarding the accuracy of the explanation), and currency (explanation should contain recent information). Moreover, along with the automatic representation of the explanation, the users can request additional information if needed. We have also described five dimensions of XAI effects: trust, transparency, understandability, usability, and fairness. We investigated current knowledge from selected articles to problematize future research agendas as research questions along with possible research paths. Consequently, a comprehensive framework of XAI and its possible effects on user behavior has been developed.

## 1. Introduction

Recently, the adoption and use of artificial intelligence (AI)-based applications by various business organizations have been increasing to aid decision-making. For example, the International Data Corporation (IDC) has estimated that the worldwide AI expenditure is supposed to increase to 110 billion US dollars by the end of 2024 (Adadi and Berrada, 2018; IDC, 2018). Because AI has become more prevalent, it has become routine to rely on it to make decisions in our daily lives (Stahl et al., 2021; Mahmud et al., 2022a,b). We use various intelligent systems every day, such as in content and product recommendation (Benbasat and Wang, 2005; Gruetzemacher et al., 2021; Wang et al., 2014; Choi et al., 2012), news websites, social media (Feng et al., 2020), healthcare (Haque et al., 2020), and other public services (Hengstler et al., 2016; Haque et al., 2021; Du and Xie, 2021); however, the working principle of AI systems is unclear as the machine-learning models used in different AI systems do not reveal enough information about the process through which the conclusion is derived (Castelvecchi, 2016). Furthermore, the deep neural network (DNN) models used in advanced AI systems are extraordinarily complex to explain. Only specific people who design the algorithms understand how the system works (Angelov and Soares, 2020). The opacity of AI systems can reduce end users' trust and reliance on using AI-based systems while making critical decisions (Hasan et al., 2021; Baum et al., 2011). To address this problem, researchers and practitioners have called for the requirement to provide explainable Artificial Intelligence (XAI) that allows end users to perceive the underlying working principle of the decision-making procedure (Laato et al., 2022; Tiainen, 2021). Understanding the working principles of AI systems is crucial for end users to make effective decisions in different contexts (Scott et al., 1977). For example, in mission-critical use cases, such as healthcare, the decision-making procedure should be understandable for the users (doctors) to rely on the system (Lauritsen et al., 2020).

\* Corresponding author.
  *E-mail addresses:* bahaul.haque@lut.fi (A.B. Haque), najmul.islam@lut.fi (A.K.M.N. Islam), patrick.mikalef@ntnu.no, patrick.mikalef@sintef.no (P. Mikalef).

Furthermore, the General Data Protection Regulation (GDPR) has also emphasized the explainability of AI systems by introducing the "right to explanation" (Goodman and Flaxman, 2017). The regulation includes another policy related to "automated individual decision-making, including profiling," to prevent personal data from being used and processed by automated systems without permission (Malgieri, 2019). In addition, the High-level Expert Group on Artificial Intelligence of the European Commission has also outlined the importance of an explanation to achieve the transparency and reliability of AI in their "Ethics Guidelines for Trustworthy Artificial Intelligence (AI)".[1] Furthermore, governments worldwide are currently adopting automated decision making; for example, the Dutch immigration services are testing automated processes for asylum requests and resident permit applications (Janssen et al., 2020). Such sensitive decision making by government organizations should have explainability for the users as well for those involved in decision making. Therefore, organizations involved in the government should also have XAI as a prerequisite for an automated decision-making system. AI-based decision-making systems can be used for scalable and larger ecosystems; however, the systems need to be incorporated with some principles related to ethics and rights (Fjeld et al., 2020).

Therefore, due to the wide applicability and demand, researchers have investigated XAI across various domains and perspectives. Previously published Systematic Literature Reviews (SLRs) on XAI have focused on the ethical perspective of AI's black box nature (Meske et al., 2022; Wells and Bednarz, 2021), human-centric design patterns for ML-based systems (Chromik and Butz, 2021), personalized explanations of XAI (Schneider and Handali, 2019), behavioral interactions of human and autonomous agents (Anjomshoae et al., 2019), XAI in healthcare domain (Chakrobartty and El-Gayar, 2021; Antoniadi et al., 2021), and AI system communication, design recommendations and tradeoffs of the end user-centric AI (Laato et al., 2022), among others. Despite the plethora of these types of investigations, we have identified two major research gaps concerning end users' explanation needs. First, most prior SLRs (see Table 1) focused on a single domain (e.g., healthcare, transportation, etc.). This limits our understanding of how the end users' explanation needs might vary across different domains. For example, the healthcare professionals' explanation needs for making critical decisions would be significantly different than consumers' decisions regarding their next purchases. Second, most prior SLRs (see Table 1) have been conducted from a technical perspective. To understand the explanation needs of end users, an SLR that reviews studies from the human perspective is needed. However, very few SLR studies (Laato et al., 2022) have been done from the end users' perspectives. Therefore, an SLR conducted across different domains that includes the latest published articles can provide a comprehensive outline of how XAI has advanced in different application domains in recent times. Moreover, a comprehensive study of human-centered XAI can help researchers and practitioners understand how people perceive different types of explanations provided by AI-based systems. The analysis will also provide meticulous insight into the impact of XAI on humans. Hence, we conducted an SLR to critically analyze the previous research on AI users' explanation needs to fulfill the research objectives, which are (1) a synthesis of prior literature on XAI that contains (a) a critical analysis of extant literature to represent current knowledge on XAI in terms of explanation needs and XAI effects and (b) research domains and (2) the development of thematically organized future research avenues.

To address the research objectives, 58 publications were selected by scanning Scopus and Web of Science databases and using rigorous citation chaining techniques. Our SLR has three key findings. First, we found four dimensions of end users' explanation needs: format, completeness, accuracy, and currency. We then linked these dimensions with the five effects of XAI: trust, transparency, understandability, usability, and fairness, which have been discussed in prior literature (Laato et al., 2022) to develop a framework. Finally, we found 10 application domains where XAI research has been conducted. Based on these findings, our paper contributes to the existing XAI literature (Binns et al., 2018; Chazette and Schneider, 2020; Schneider et al., 2021; van der Waa et al., 2020; Laato et al., 2022) by 1) identifying the dimensions of end users' explanation needs and presenting them from information systems research perspective; 2) identifying the outcomes of XAI from the end users' perspectives; 3) identifying research gaps and problematizing future research directions in XAI, particularly from end users' perspectives; and 4) building a framework for XAI research from end users' perspective. Our findings also help practitioners design a more user-friendly and trustworthy XAI system by determining the explanation needs of the end users.

The remainder of the paper is structured as follows. Section 2 outlines the background of XAI and Related Works. Section 3 describes the SLR methodology and literature selection. Section 4 contains the research trend of XAI based on the selected articles, and Section 5 synthesizes previous studies on XAI. This section comprehensively represents the current knowledge of XAI aligned with information systems research. Section 6 critically analyzes the current knowledge to identify future research directions. Section 7 outlines the comprehensive framework of XAI research from the end user's perspective. Section 8 briefly describes the implications of this work, and Section 9 concludes the paper.

## 2. Background

### 2.1. Explainable AI and related concepts

Explainable AI, interpretable AI, transparent AI, understandable AI, and responsible AI terminologies are used interchangeably in the literature (Arrieta et al., 2020). XAI has emerged intending to present explanations purveyed to human understanding, trust, and transparency (Gerlings et al., 2021a, 2021b). The relational link connecting the input and the output of an artificial neural network is not observable. Therefore it is necessary to put effort into the explainability and interpretability of the black-box nature of various AI models (Dağlarli, 2020). DARPA, one of the leading research organizations on XAI, explained XAI as an extension of an AI system whose models and decisions can be easily understandable and properly believable by end users (Gunning and Aha, 2019). The understandability and believability of machine-learning models contribute to the interpretability of a machine-learning model for the target audience (Lipton, 2018). Explainability usually indicates how strongly a particular phenomenon can be described so that the audience can effortlessly understand it. Therefore, in XAI, explainability means the AI should be capable of explaining predictions obtained from a model from a more profound methodological point of view to users (Antunes et al., 2012); however, explainable AI can also be defined as: "given an audience, an explainable Artificial Intelligence produces details or reasons to make its functioning clear or easy to understand" (Arrieta et al., 2020).

Interpretability (Lipton, 2018) specifies that the working procedure of the machine models should be made unambiguous and crystal clear to both technical and nontechnical users. Though interpretability and explainability are used interchangeably, there are some basic conceptual differences between them. Explainability means explaining the

---

[1] on Artificial Intelligence (AI HLEG), H.-L. I. G. (2019). *Ethics Guidelines for Trustworthy AI.* European Commission. https://ec.europa.eu/digital-single-market/en/news/ethics-guidelines-trustworthy-ai

**Table 1**
Comparison of previous related systematic review studies.

| Study | Purpose | Years included | Source of primary studies |
|---|---|---|---|
| Anjomshoae et al., 2019 | Presents a goal-driven literature review of explainable robots and agents to enhance the understanding of the "black box." | All documents are published between the years 2008–2018. | An initial collection of 303 papers were reduced to 62 final selections using seven inclusion criteria. The authors did not mention the types of individual publication. These papers were collected from digital libraries, such as IEEExplore, Science Direct, ACM, and Google Scholar. |
| Schneider and Handali, 2019 | This study provides a structured collection of information that conceptualizes "personalized explanation" and relates the idea to other domains that are intertwined with XAI. | The paper did not mention the publication time of these documents. | They collected research articles and conference papers from the IEEE Xplore, AIS, ACM, and Arxiv databases. Their study did not mention the total number of papers considered. |
| Antoniadi et al., 2021 | Highlighting the indispensability of interpretable AI systems in medical use cases, this study underscores the ethical and fair decision making by AI systems in medical practices. This study claims to provide suggestions to aid future opportunities and tackle foreseeable challenges. | Unknown – 2020 (authors did not specify their starting year as a search criterion). | Using the Google Scholar database, they identified 668 articles based on six combinations of search phrases. Through an intricate elimination and selection process, 33 papers were finally selected. The authors did not specify the publication type of these papers. |
| Chakrobartty and El-Gayar, 2021 | Raising concerns about the un-explainability of AI techniques, especially in the medical sector, this study highlights the methods and practices that emphasize XAI in the medical sector. | This study covers documents published between 2008 and 2020. | Based on eight search keywords, they initially found 66 documents, which were reduced to 22 using several inclusion and exclusion criteria. The authors did not specify the type of publications. |
| Chromik et al., 2021 | To better comprehend the black box, this study presents an argument that advocates that the explanation user interfaces interpretability increases by employing explanation-generating models. This study provides insight into how designers can attune the explanation of AI systems in user interfaces. | Unknown – 2020 (authors did not specify their starting year as a search criteria). | An initial collection of 146 documents was reduced to 91 documents that meticulously matched the research objective. |
| Gerlings et al., 2021b | This study presents a thoroughgoing discussion on how XAI addresses the black box problem in AI-based applications. By conducting a comprehensive study of recent publications, they attempted to find how XAI contributes to reducing the gap between stakeholders and the black box. | Covers documents from 2016 to 2020. | They collected data from ArXiv, AIS, JSTOR, ACM Digital Library, IEEE Xplore, SAGE, and Science Direct digital libraries. From 221 initial documents, they finally picked 64 documents for their study. |
| Linardatos et al., 2021 | This study highlights the programming implementation in recent studies that contributes to increasing the interpretability of ML models from both theorist and practitioner perspectives. | Not specified. | Not specified. |
| Wells & Bednarz, 2021 | This study accentuates the societal and ethical implications of XAI in the area of reinforcement learning. The study showed limitations, such as lack of user studies, the prevalence of toy-examples, and difficulties providing understandable explanations, in the case of reinforcement learning. | Covers published documents between 2014 and 2020. | Conducting a Boolean search on digital libraries, such as ACM, IEEExplorer, Science Direct, and Springer Link digital libraries, they gathered 520 papers, among which they justify choosing only 25 papers that matched their research interest.<br>• Defense/Military – 1<br>• Autonomous Vehicles – 2<br>• Networking 2<br>• Robotics – 4<br>• Gridworld – 5<br>• Games - 16 |
| Laato et al., 2022 | The authors identified the high-level objectives of AI communications with end users such as understandability, trustworthiness, transparency, controllability, and fairness. Moreover, the authors provide design recommendations for explanations of AI systems. | Search conducted on October 2020 | The search was conducted on both Scopus and Web of Science from XAI from the HCI perspective. 808 unique articles were extracted after removing the duplicates. The final sample size was 25 articles that matched their research objective. |
| Our study | Our study focuses on<br>(1) a synthesis of prior literature on XAI that contains critical analysis of extant literature to represent current knowledge on XAI in terms of explanation representation, XAI effects, explored research domains and<br>(2) the development of thematically organized future research avenues. | Covers published documents up to July 15, 2022. | Conducted a search on Scopus and Web of Science, which are the two most comprehensive databases for scholarly publications. Based on a wide range of keywords, the search initially revealed 2896 studies both from the Scopus and Web of Science. From them, 58 studies were included as they matched the research objectives and interests. Only end user-centric empirical studies are included, which provides a unique identifier for our study.<br>In contrast to other SLRs (e.g., Laato et al., 2022), we have discussed the explanation representation across different domains, have identified explanation quality dimensions through Wixom and Todd (2005), and have analyzed the effect of XAI across different domains and categorically presented them. Finally, we have provided a synthesized framework by connecting explanation dimensions and XAI effects. |

decisions made by machine models in a human-understandable form, and interpretability is the explanation of how or why a model resulted in a particular prediction (Doshi-Velez and Kim, 2017). Transparency is another critical mainstay of XAI, which means being effortlessly viewed through something. In XAI, a model can be considered transparent if it can explain its different steps simplistically to human users (Wachter et al., 2017). Understandability specifies whether the features and attributes of a model are easily recognizable by users without knowing its inner composition. Understandable AI specifies whether the unification of model developers and UI designers can produce a human-centric AI architecture (Arrieta et al., 2020). Finally, responsible AI is a framework from the governance perspective that is comprised of guidelines and policies for AI technologies to ensure integrity, efficiency, and productivity. These policies and guidelines also involve responsible system design, proper monitoring, and awareness (Ghallab, 2019).

*2.2. Related works*

Practical relevance and research interest in XAI have significantly increased in recent times. We have been able to identify several prior SLRs which focus on various domains. Table 1 represents a comparative analysis of these identified studies. The black-box nature of AI poses ethical concerns and risks since no one can interpret what is going on inside and how the data is being processed (Meske et al., 2022). Therefore, the open development of AI should be closely observed and audited, as the compromises involved may lead to dire consequences (Meske et al., 2022). The explanatory design of the user interface can also contribute to understanding black box AI. Interaction factors, such as transmission, dialogue, control, experience, optimal behavior, tool use, and embodied action, are critical when designing such a system. Four human-centric design patterns for ML-based systems increase the understanding level of a human user through a set of explanation-generating methods (Chromik and Butz, 2021). Naturalness, responsiveness, flexibility, and sensitivity are the four recurring design patterns that are the most frequently used human-centric design patterns (Chromik and Butz, 2021). Personalized explanation enhances the interpretability and understanding of the explainees (Schneider and Handali, 2019); however, there is a substantial research gap regarding collecting personalized and explicit information from the explainees with arguable privacy concerns (Schneider and Handali, 2019).

Research on explainable AI has increased and has primarily focused on policy summarization, human collaboration, visualization, verification, etc. (Wells and Bednarz, 2021); however, research gaps exist in customized algorithms, user testing, and scalability (Wells and Bednarz, 2021). In one systematic review, the explainable nature of the behavioral interaction of agents and robots with human users is discussed (Anjomshoae et al., 2019). The work also summarizes the importance of the explainable nature of intelligent systems for non-expert users. Both technical and non-technical perspectives are important for the XAI domain. One of the seminal scholarly works (discussing the importance of unveiling the black box) comprehensively outlines the need, research challenges, and future research opportunities to provide explainability (Gerlings et al., 2021a, 2021b). Healthcare is a crucial domain for any type of technology use. Similarly, XAI research for the healthcare domain can help doctors make decisions (Chakrobartty and El-Gayar, 2021). XAI in healthcare includes various techniques and methods used for XAI (Chakrobartty and El-Gayar, 2021) and clinical decision making (Antoniadi et al., 2021). Other traditional review studies discuss explainable AI from a technical perspective, such as the interpretability methods of various machine-learning interpretability models (Linardatos et al., 2021). Recently, Laato et al. (2022) identified the high-level objectives of AI communications with end users such as understandability, trustworthiness, transparency, controllability, and fairness. Moreover, they provided design recommendations for explanations of AI systems along with future research directions.

## 3. Methodology

We have adopted an SLR methodology to summarize the existing studies on XAI. An SLR is a method for locating, assessing, and evaluating relevant research for certain research questions, topics, or phenomena being studied (Kitchenham and Charters, 2007). To analyze prior contributions to AI, the techniques focus on "identifying, evaluating and interpreting all available research relevant to a particular research question, or topic area, or phenomenon of interest" (Kitchenham and Charters, 2007). We identified the search terms and used Boolean operators to generate search strings for searching the Scopus and Web of Science databases. Scopus and Web of Science are among the most comprehensive and recognized databases with reputed scholarly publications. We did not specify any starting year for the search criteria, and therefore the timeline used for the search results would be the day we conducted our last search query, which is July 15th, 2022. We have identified inclusion and inclusion criteria to filter irrelevant studies and to develop the final article list. The search terms are shown in Table 1.

*3.1. Literature selection criteria*

For the literature selection, we defined a set of well-defined inclusion and exclusion criteria based on the scope of this review work. The inclusion and exclusion criteria are outlined in Table 2.

*3.1.1. Search result extraction and analysis*

The search terms and the results extracted are provided in Table 3.

From both databases, only conference and journal articles were selected, and the duplicates were removed, leaving 1707 articles. After reading the titles and abstracts, 1190 articles were removed from the list. Full texts of the remaining 517 articles were studied carefully to remove the articles that were not within the scope of our research theme. Furthermore, articles without empirical studies were excluded, which resulted in the final 58 articles. Fig. 1 depicts the screening and selection process.

## 4. Research trend

Of the 58 studies in our SLR, 13 were journal articles, and 45 were conference articles. Table 4 depicts the publications per year for the selected studies and the number of journal and conference articles. Here, we observed that the number of publications increased from 2018 onwards. This clarifies that explainable AI has been a topic of interest in recent years. Other bibliometric data of the selected articles, such as the number of publications by publishers (Table 5) and the top-five cited articles (according to Scopus), including their author affiliation (Table 6), are presented as well.

## 5. Synthesis of prior literature

This section provides a critical analysis of the selected research studies and an overview of their findings. This section is divided into (1) Current Knowledge Representation and (2) Research Domains. Table 7 represents the synthesis of prior literature.

*5.1. Current knowledge representations*

This section represents the current knowledge extracted from the selected articles. The section is divided into three subsections: (1) XAI representation, (2) Effects of Explainable AI, and (3) Explanation Presentation Time.

*5.1.1. XAI representation*
We have adopted the information quality dimensions proposed by Wixom and Todd (2005) to conceptualize XAI representation dimensions. Wixom and Todd (2005) proposed four information quality dimensions. The dimensions are "format," which defines the user's perception regarding the information presented; "completeness," which represents how adequately all the necessary information is presented; "accuracy," which represents the user's perception of how accurate the information is; and "currency," which represents the user's perception of how up-to-date the information is at the time of access. To understand the quality of the explanation presented by the XAI system, we have aligned our findings with these four dimensions (Wixom and Todd, 2005). These dimensions are described based on the data extracted from the literature.

**Table 2**
Inclusion and exclusion criteria.

| Inclusion criteria | Exclusion criteria |
| --- | --- |
| 1. Conference and journal papers are included | 1. Review articles, book chapters, magazines, and editorials were excluded |
| 2. Only publications in English language | 2. Articles published other than in the English language were excluded |
| 3. Full-text availability in online databases and repositories | 3. Full-text was not available on online repositories |
| 4. Articles with empirical studies are included (end user centric) | 4. Duplicate results were removed |

*5.1.1.1. Format.* The information representation of XAI systems is primarily either textual, visual, or auditory or in a hybrid mode. Users of different domains have different perceptions of the information representation format. The users of healthcare-related AI systems need explanations in both textual and visual(graphical) formats (Branley-Bell et al., 2020; Cheng et al., 2019; Daudt et al., 2021; Lee and Rich, 2021; Wang et al., 2019; Xie et al., 2019; Rodriguez-Sampaio et al., 2022). For example, if an explanation is presented with appropriate figures, images, and terminologies, the user's understandability increases (Bussone et al., 2015; Cai et al., 2019; Eiband et al., 2019; Hudon et al., 2021). The hybrid explanation format reveals important information about the reasoning involved in a decision-making process (Branley-Bell et al., 2020; Górski and Ramakrishna, 2021). Expert end users of AI-based diagnostic pathology prefer a user-centric design that combines textual and visual explanations (Evans et al., 2022). Users of music recommendations, movie recommendations, drawing tools, and other different types of sports-related systems need explanations in a hybrid format to increase understandability (Cramer et al., 2008; Ehsan et al., 2019; Kouki et al., 2019; Ngo et al., 2020; Oh et al., 2018; Schmidt et al., 2020; Szymanski et al., 2021). The hybrid explanation can include a partial dependence plot as well as documentation. This two-dimensional plot describes how one output is influenced by another input (Szymanski et al., 2021).

Users prefer hybrid explanations that include textual and visual explanations and, in some cases, explanations using indicator lights (Schneider et al., 2021; van der Waa et al., 2020). Explanations in e-commerce (Ehsan et al., 2021; Eslami et al., 2018), education (Cheng et al., 2019; Conati et al., 2021; Mucha et al., 2021; Putnam and Conati, 2019), finance (Binns et al., 2018; Chromik et al., 2021; Cirqueira et al., 2020), law (Liu et al., 2021a, 2021b; Górski and Ramakrishna, 2021),

**Table 3**
Search terms and results.

| Search terms | Databases | Results |
| --- | --- | --- |
| "Explainable Artificial Intelligence" OR "Transparent Artificial Intelligence" OR "Interpretable Artificial Intelligence" OR "Understandable Artificial Intelligence" OR "Artificial Intelligence Transparency" OR "Artificial Intelligence Interpretability" OR "Artificial Intelligence Understandability" OR "XAI" OR "Understandable AI" OR "AI Transparency" OR "AI Interpretability" OR "Responsible AI" OR "AI Decision Making" OR "AI trust" OR "AI system use" OR "AI use" | Scopus Web of Science | 2669 1345 |

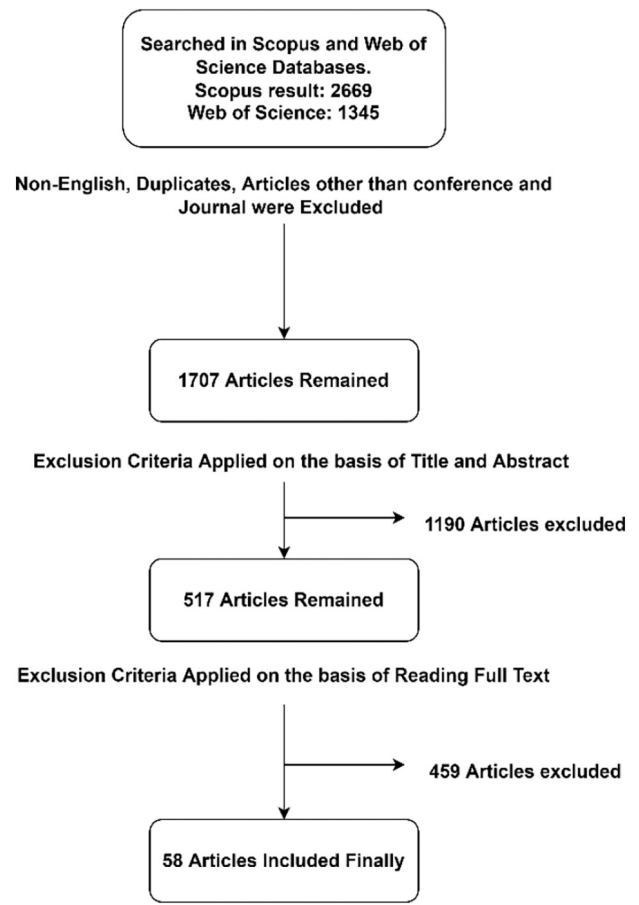

**Fig. 1.** Article screening and selection process.

**Table 4**
Number of conference and journal publications by year.

| Publication year | Conference publications | Journal publications |
| --- | --- | --- |
| 2008 | 0 | 1 |
| 2009 | 2 | 0 |
| 2010 | 1 | 0 |
| 2015 | 1 | 0 |
| 2018 | 4 | 0 |
| 2019 | 11 | 0 |
| 2020 | 10 | 4 |
| 2021 | 14 | 4 |
| 2022 | 2 | 4 |

**Table 5**
Number of articles by publisher.

| Publishers | No. of publication |
| --- | --- |
| Taylor and Francis | 1 |
| Springer | 12 |
| SAGE | 1 |
| IEEE | 2 |
| Emerald | 1 |
| Elsevier | 4 |
| ACM | 37 |

and social networking (Lim and Dey, 2009; Yin et al., 2019) can also be provided in the hybrid mode. For example, the logic behind algorithmic decision-making, working procedures, and product image in e-commerce systems are all discussed in the textual explanation. For law-related systems, counterfactual explanations and logical reasoning behind the decision in laymen's terms are needed (Górski and

**Table 6**
Top-five cited articles, including their authors and affiliations according to Google scholar (till the date of final submission of this article).

| Title | Year | Source | Cited by | Authors with affiliations | Publisher | Type |
|---|---|---|---|---|---|---|
| Designing theory-driven user-centric explainable AI | 2019 | Conference on Human Factors in Computing Systems - Proceedings | 447 | Wang, D., School of Computing, National University of Singapore, Singapore, Singapore; Yang, Q., Human-Computer Interaction Institute, Carnegie Mellon University, Pittsburgh, PA, United States; Abdul, A., School of Computing, National University of Singapore, Singapore, Singapore; Lim, B.Y., School of Computing, National University of Singapore, Singapore, Singapore | ACM | Conference |
| It's reducing a human being to a percentage"; perceptions of justice in algorithmic decisions | 2018 | Conference on Human Factors in Computing Systems - Proceedings | 353 | Binns, R., Dept. of Computer Science, University of Oxford, United Kingdom; Van Kleek, M., Dept. of Computer Science, University of Oxford, United Kingdom; Veale, M., Dept. of Science, Technology, Engineering and Public Policy, University College London, United Kingdom; Lyngs, U., Dept. of Computer Science, University of Oxford, United Kingdom; Zhao, J., Dept. of Computer Science, University of Oxford, United Kingdom; Shadbolt, N., Dept. of Computer Science, University of Oxford, United Kingdom | ACM | Conference |
| Why and why not explanations improve the intelligibility of context-aware intelligent systems | 2009 | Conference on Human Factors in Computing Systems - Proceedings | 568 | Lim, B.Y., Carnegie Mellon University, United States; Dey, A.K., Carnegie Mellon University, 5 United States; Avrahami, D., Intel Research Seattle, United States | ACM | Conference |
| Assessing demand for intelligibility in context-aware applications | 2009 | ACM International Conference Proceeding Series | 243 | Lim, B.Y., Carnegie Mellon University, Pittsburgh, United States; Dey, A.K., Carnegie Mellon University, United States | ACM | Conference |
| The effects of transparency on trust in and acceptance of a content-based art recommender | 2008 | User Modeling and User-Adapted Interaction | 435 | Cramer, H., Human Computer Studies Lab., University of Amsterdam, Amsterdam, Netherlands; Evers, V., Human Computer Studies Lab., University of Amsterdam, Amsterdam, Netherlands; Ramlal, S., Human Computer Studies Lab., University of Amsterdam, Amsterdam, Netherlands; Van Someren, M., Human Computer Studies Lab., University of Amsterdam, Amsterdam, Netherlands; Rutledge, L., Telematica Institute, Enschede, Netherlands, CWI, Amsterdam, Netherlands; Stash, N., Eindhoven University of Technology, Eindhoven, Netherlands, VU University Amsterdam, De Boelelaan Amsterdam, Netherlands; Aroyo, L., Eindhoven University of Technology, Eindhoven, Netherlands, VU University Amsterdam, Amsterdam, Netherlands; Wielinga, B., Human Computer Studies Lab., University of Amsterdam, Amsterdam, Netherlands | Springer | Journal |

Ramakrishna, 2021). The explanation format in virtual assistant systems includes voice-based interactions along with textual and visual explanations (Weitz et al., 2019, 2021; Gao et al., 2022). An interactive agent with hybrid (textual and audio-visual) explanations can increase the perception of trust in a system (Weitz et al., 2021). XAI in immigration systems needs both the textual and visual information format because the decision-making requires careful observation of personal details, travel itineraries, and photo matches with travelers (Janssen et al., 2020). In the human resource context, both textual and visual explanations are recommended (Bankins et al., 2022). For criminal justice use cases, the reasoning should include information related to both "why" and "why not" because the counterfactual details help clear any doubt or bias (Dodge et al., 2019). The hybrid explanation format is also required for other context-aware systems (Lim et al., 2009), general decision-making systems (Brennen, 2020; Schrills and Franke, 2020), travel guides (Lim and Dey, 2009), cooking recommendation systems (Broekens et al., 2010), and wearable systems (Danry et al., 2020).

*5.1.1.2. Completeness.* Completeness in XAI refers to providing the target user with all required information, including on demand supplementary data. For the healthcare domain, the user needs to be presented with patients' demographic information, cardinal symptoms, previous test data, and initial evaluations (Wang et al., 2019; Xie et al., 2019). The visual explanation can include a vivid and concise representation of appropriate diagnosis images, indicators of different properties, bar charts, etc. (Bussone et al., 2015; Cai et al., 2019; Ehsan et al., 2019; Eiband et al., 2018). The textual explanation can include a detailed representation of the decision-making procedure and the algorithms' working principles (Branley-Bell et al., 2020; Bussone et al., 2015; Cai et al., 2019; Daudt et al., 2021; Eiband et al., 2019; Lee and Rich, 2021). In addition, providing users with contextual information and references about the prediction upon request increases users' trust and perception of reliability (Branley-Bell et al., 2020; Daudt et al., 2021; Lee and Rich, 2021; Bove et al., 2021). Contextual information refers to an explanation that is domain-specific or application-specific. Along with explaining the algorithms and machine-learning models, it is also important to include domain-specific contextual information regarding decision making. The contextual information varies across different domains. Therefore, it should be considered by the developers during the design phase of a system. Media and entertainment recommendation systems can explain decision-making by revealing the working procedure of the algorithm, the personal data being used, and visual representation of the recommendation being made (Ehsan et al., 2019; Kouki et al., 2019; Schmidt et al., 2020). For example, the users of music and movie recommendation systems want to see what kind of data has been used for the prediction and the popularity rating of the decision (Kouki et al., 2019; Ngo et al., 2020). In addition, information regarding the movie name, previous ratings, genres, and confidence measurements can be provided as an explanation.

Another example is an online news recommendation system, where the visual explanation includes a two-dimensional partial dependence plot that describes how the output is influenced by the input properties (Szymanski et al., 2021). A textual explanation of XAI can also include product type, price, order details, and other different attributes and features (Ehsan et al., 2021; Eslami et al., 2018; Bankins et al., 2022). The reasons for user agreement and disagreement related to predictions

**Table 7**
Synthesis of prior literature.

| Source | XAI representation | Effects | Explanation presentation time | Research focus |
| --- | --- | --- | --- | --- |
| Cramer et al., 2008 | Hybrid representation | Accuracy, Trust Transparency | With the recommendation and after the user demands explanation | Media and Entertainment |
| Branley-Bell et al., 2020 | Hybrid representation | Trust Understandability | With the recommendation and after the user demands explanation | Healthcare |
| Cheng et al., 2019 | Hybrid representation | Understandability | After the user demands explanation as supplementary information | Education |
| Daudt et al., 2021 | Hybrid representation | Trust Understandability | Not mentioned explicitly, but analysis shows both with recommendation and after the user demands explation | Healthcare |
| Lee and Rich, 2021 | Hybrid representation | Trust | With the recommendation and after the user demands explanation | Healthcare |
| Wang et al., 2019 | Hybrid representation | Trust Understandability | Not mentioned explicitly | Healthcare |
| Xie et al., 2019 | Hybrid representation | Trust | Not mentioned explicitly | Healthcare |
| Rodriguez-Sampaio et al., 2022 | Hybrid representation | Trust Understandability | With the recommendation and after the user demands explanation | Healthcare |
| Bussone et al., 2015 | Graphical representation | Trust | With the recommendation and after the user demands explanation | Healthcare |
| Cai et al., 2019 | Graphical representation | Transparency | With the recommendation and after the user demands explanation | Healthcare |
| Eiband et al., 2019 | Graphical representation | Transparency | With the recommendation and after the user demands explanation | Recommendation System |
| Hudon et al., 2021 | Hybrid representation | Trust Understandability | Not explicitly mentioned | Media and Entertainment |
| Górski and Ramakrishna, 2021 | Hybrid representation | Understandability Fairness | With the recommendation and after the user demands explanation | Law |
| Evans et al., 2022 | Hybrid representation | Understandability | With the recommendation and after the user demands explanation | Healthcare |
| Ehsan et al., 2019 | Hybrid representation | Understandability | With the recommendation | Media and entertainment |
| Kouki et al., 2019 | Hybrid representation | Trust | With the recommendation | Media and entertainment |
| Ngo et al., 2020 | Hybrid representation | Transparency | With the recommendation | Media and entertainment |
| Oh et al., 2018 | Hybrid representation | Trust Usability | With the recommendation and after the user demands explanation | Media and entertainment |
| Schmidt et al., 2020 | Hybrid representation | Trust | With the recommendation and after the user demands explanation | Media and entertainment |
| Szymanski et al., 2021 | Hybrid representation | Understandability Transparency | With the recommendation and after the user demands explanation | Media and entertainment |
| Ehsan et al., 2021 | Hybrid representation | Trust Understandability | With the recommendation and after the user demands explanation | E-commerce |
| Eslami et al., 2018 | Hybrid representation | Understandability | With the recommendation and after the user demands explanation | E-commerce |
| Conati et al., 2021 | Hybrid representation | Transparency | With the recommendation and after the user demands explanation | Education |
| Mucha et al., 2021 | Hybrid representation | Fairness | With the recommendation and after the user demands explanation | Education |
| Putnam and Conati, 2019 | Hybrid representation | Trust | With the recommendation and after the user demands explanation | Education |
| Li et al., 2021 | Hybrid representation | Transparency | With the recommendation and after the user demands explanation | Human Resource Management |
| Khosravi et al., 2022 | Hybrid representation | Trust | With the recommendation and after the user demands explanation | Education |
| Binns et al., 2018 | Hybrid representation | Understandability Fairness | With the recommendation | Transportation, Finance |
| Chromik et al., 2021 | Hybrid representation | Trust Understandability Usability | With the recommendation and after the user demands explanation | Finance |
| Cirqueira et al., 2020 | Hybrid representation | Trust | With the recommendation and after the user demands explanation | Finance |
| Liu et al., 2021a | Hybrid representation | Transparency Understandability Fairness | With the recommendation and after the user demands explanation | Legal |
| Liu et al., 2021b | Hybrid representation | Transparency | With the recommendation and after the user demands explanation | Social Networking |
| Górski and Ramakrishna, 2021 | Hybrid representation | Transparency Understandability Fairness | With the recommendation and after the user demands explanation | Legal |
| Lim and Dey, 2009 | Hybrid representation | Understandability | With the recommendation and after the user demands explanation | Social Networking |
| Yin et al., 2019 | Hybrid representation | Trust | With the recommendation and after the user demands explanation | Social Networking |
| Weitz et al., 2019 | | Trust | With the recommendation and after the user demands explanation | Digital Assistant |

**Table 7** (*continued*)

| Source | XAI representation | Effects | Explanation presentation time | Research focus |
| --- | --- | --- | --- | --- |
| Weitz et al., 2021 | Hybrid representation | Trust | With the recommendation and after the user demands explanation | Digital Assistant |
| Janssen et al., 2020 | Hybrid representation | Transparency Fairness | With the recommendation and after the user demands explanation | E-Governance |
| Bankins et al., 2022 | Hybrid representation | Trust | With the recommendation | Human Resource Management |
| Dodge et al., 2019 | Hybrid representation | Trust Fairness | With the recommendation and after the user demands explanation | E-Governance |
| Lim et al., 2009 | Hybrid representation | Understandability | With the recommendation and after the user demands explanation | Recommendation System |
| Brennen, 2020 | Hybrid representation | Trust Usability | With the recommendation and after the user demands explanation | Recommendation System |
| Schrills and Franke, 2020 | Hybrid representation | Trust | With the recommendation and after the user demands explanation | Recommendation System |
| Broekens et al., 2010 | Hybrid representation | Understandability | With the recommendation and after the user demands explanation | Media and Entertainment |
| Danry et al., 2020 | Hybrid representation | Understandability | With the recommendation and after the user demands explanation | Healthcare |
| Eiband et al., 2018 | Graphical representation | Transparency | With the recommendation and after the user demands explanation | Healthcare |
| Bove et al., 2021 | Graphical representation | Trust Understandability | With the recommendation and after the user demands explanation | E-commerce |
| Chazette and Schneider, 2020 | Hybrid representation | Understandability | With the recommendation | Transportation |
| Schneider et al., 2021 | Hybrid representation | Understandability | With the recommendation | Transportation |
| van der Waa et al., 2020 | Hybrid Representation | Understandability Transparency | With the recommendation | Transportation |
| Park et al., 2021 | Hybrid representation | Trust | With the recommendation and after the user demands explanation | Human Resource |
| Hong et al., 2020 | Hybrid representation | Trust | With the recommendation and after the user demands explanation | Social networking |
| Liao et al., 2020 | Hybrid representation | Trust Transparency | With the recommendation and after the user demands explanation | Social networking |
| Wang and Moulden, 2021 | Hybrid representation | Transparency | With the recommendation and after the user demands explanation | Social networking |
| Dhanorkar et al., 2021 | Hybrid representation | Trust | With the recommendation and after the user demands explanation | AI Development |
| Evans et al., 2022 | Hybrid representation | Trust | With the recommendation and after the user demands explanation | Healthcare |
| Andres et al., 2020 | Hybrid representation | Trust | On demand explanation | AI Development |
| Hind et al., 2020 | Hybrid representation | Understandability | With the recommendation and after the user demands explanation | Social networking |

in intelligent tutoring systems must be explained to the users as well (Conati et al., 2021; Putnam and Conati, 2019). Therefore, users should be able to request information if the explanations provided do not meet their expectations. Furthermore, other systems, such as grade estimations and university admission decision making, are required to provide students with personal details, academic details, and other required attributes that contribute to decision making (Cheng et al., 2019; Mucha et al., 2021).

Loan application systems, fraud detection, and other banking software are sophisticated decision-making systems. Therefore, explanations for these types of systems should be more detailed and comprise personal details, previous credit history, employment history, and the algorithms' working procedures (Binns et al., 2018; Chromik et al., 2021; Cirqueira et al., 2020). Similarly, for transportation systems, decision making can be explainable using contextual information, confidence measurements, light indicators, and previous decisions in similar situations (Chazette and Schneider, 2020; Schneider et al., 2021; Bove et al., 2021). Flight re-routing systems can provide the reasoning behind choosing specific routes and other supplementary information on demand (Binns et al., 2018). Virtual assistant systems should provide an explanation using the appearance of a virtual agent, such as facial expressions, voice, and gestures. A harmonic combination of explainable AI methods as well as appropriate linguistic representations can make the system trustworthy (Weitz et al., 2019, 2021; Gao et al., 2022). XAI in a human resource management system should explain the working procedure and should display personal information and other attributes both in a textual and visual format (Park et al., 2021). A similar situation is observed in the case of immigration services and criminal justice use cases (Dodge et al., 2019; Janssen et al., 2020). To establish the completeness of the XAI system, the system developers and designers should keep a critical eye on the explanation types and user requirements. The users want "why," "why not," "how," "what if," and "what else" explanations from the systems along with an interactive user interface (Broekens et al., 2010; Conati et al., 2021; Schrills and Franke, 2020). Moreover, developers may consider using different color-coding indicators that can also enhance trust among users (Brennen, 2020). Therefore, they should design and develop an interactive system carefully considering all the requirements of users, device diversity, and regulatory issues to promote the completeness of XAI (Brennen, 2020; Danry et al., 2020; Hong et al., 2020).

*5.1.1.3. Accuracy.* Users' accuracy perceptions regarding information from XAI systems vary and depend on different factors. Explanations containing personalized prioritization matrices, counterfactual information about specific predictions (Liu et al., 2021a, 2021b), and supplementary information instigate the perception of accuracy and understandability among users (van der Waa et al., 2020; Wang et al., 2019; Xie et al., 2019). Moreover, this information can help verify decision-making and motivate the user to adopt an AI-based system (Wang et al., 2019). The academic tutoring system shows the confidence value of a decision as an explanation, which helps users accept or ignore a decision (Putnam and Conati, 2019). In addition, for education-related AI tools, explanation accuracy can increase if comparisons are shown between previous and current recommendations, trust scores of different recommendations using different models, etc. (Li et al., 2021; Khosravi et al., 2022). Some explainable AI systems can increase user interaction by providing detailed user instructions (Oh et al., 2018), information of the mental model used (Cramer et al., 2008), collaborative filtering (Ngo et al., 2020), and contextual data (Eiband et al., 2018; Liao et al., 2020; Bove et al., 2021). User involvement in the design process reduces the knowledge gap and promotes accuracy perceptions (Eslami et al., 2018; Ngo et al., 2020; Oh et al., 2018). Users' accuracy perceptions of XAI information are based on an explanation that contains information related to the certainty level of prediction (Bussone et al., 2015; Eiband et al., 2019), algorithmic decision-making procedures (Eiband et al., 2019; Park et al., 2021), claims and evidence (Danry et al., 2020), and information regarding domain expert engagement in the development process (Mucha et al., 2021; Wang and Moulden, 2021).

XAI should produce a human-like explanation and should show the accuracy level of the system to make the system more interpretable and accurate (Janssen et al., 2020; Lim and Dey, 2009; Park et al., 2021). Users' perceptions of the accuracy of the data of the XAI system can be established if the explanation of the algorithmic working procedure is presented sequentially to the users. This sequential flow of actions and information will motivate the user to accept or deny the decision (Broekens et al., 2010; Conati et al., 2021). AI-based law-related decision-making systems can positively affect users' accuracy perceptions by including evidence-based-reasoning sentences, legal rule sentences, and citation sentences (Górski and Ramakrishna, 2021). Explanations including this information act as a reference to the accuracy of the decision made by the system (Górski and Ramakrishna, 2021).

*5.1.1.4. Currency.* Currency is defined as the user's perception of up-to-date information (Wixom and Todd, 2005); however, for XAI, currency unfolds differently. XAI explains the algorithmic working principle, counterfactual data, supplementary information, and contributing features (Binns et al., 2018; Chromik et al., 2021; Eiband et al., 2019). From the XAI perspective, though the users are presented with an automatic explanation, an on-demand explanation is also available. The on-demand explanation can include the most recent information about any decision (Bussone et al., 2015; Putnam and Conati, 2019; Schrills and Franke, 2020; Wang et al., 2019). Supplementary information regarding the contextual data and the latest and historical references can also be available in XAI systems (Binns et al., 2018; Branley-Bell et al., 2020; Cirqueira et al., 2020; Bove et al., 2021). When designers and developers include the users in the XAI development process, they can acquire up-to-date user requirements (Ngo et al., 2020; Oh et al., 2018).

For fraud detection, loan approval contexts, and other mission-critical systems, it is essential to present the latest information (Binns et al., 2018; Chromik et al., 2021; Cirqueira et al., 2020). Recruitment systems also should use the candidates up to date information for recruitment-related decision-making (Bankins et al., 2022; Li et al., 2021). Financial decision-making such as loan or credit approval, should use the up-to-date financial history of the person (Chromik et al., 2021; Cirqueira et al., 2020). Another use case related to the flight re-routing system offers the latest flight data to users so that travel is flexible and comfortable (Binns et al., 2018). The same goes for media and entertainment recommendation systems, where the users recommend the latest movies and music as part of the process (Kouki et al., 2019). Similarly, instant messaging applications and tour guide systems present the latest explanation data to the user (Lim and Dey, 2009; Yin et al., 2019).

*5.1.2. Effects of XAI*

Our goal in this paper is to link the XAI representation dimensions with XAI effects. Towards this goal, we adopted the XAI objectives described by Laato et al. (2022) and categorize explainable AI effects into trust, transparency, usability, understandability, and fairness. The effects are briefly explained in the following subsections based on our literature review.

*5.1.2.1. Trust.* Based on our literature review, we have observed that users' trust is affected by both the stated and the observed accuracy of the machine-learning model. For example, users' trust in a machine-learning model increases or decreases based on the information about stated and observed accuracy (Yin et al., 2019). Prior research studies have shown that providing users with contextual information, historical data, and the proper reference behind decision making enhances trust in the system, particularly in the context of healthcare and finances (Cirqueira et al., 2020; Kouki et al., 2019; Wang et al., 2019; Xie et al., 2019; Dhanorkar et al., 2021; Bove et al., 2021). Furthermore, users' perceptions of bias are reduced if explanations include input value attributes, reference data related to the prediction, and contextual information (Cirqueira et al., 2020; Daudt et al., 2021; Hong et al., 2020; Lee and Rich, 2021; Evans et al., 2022). Moreover, a high confidence level for specific predictions helps users build trust in the system (Bussone et al., 2015; Ehsan et al., 2021).

Explanation styles have a significant impact on users' trust in a system. For example, visual explanations of the input data of the machine-learning model induce a higher level of visibility, understandability, observability, and trust in the system (Schrills and Franke, 2020; Hudon et al., 2021). User trust also varies based on whether they are informed that a human or AI made the decision; however, the variation is mostly observed when the decision is positive. Users tend to trust a system if the decision is positive irrespective of the decision maker (Bankins et al., 2022). The explanation should contain enough detail regarding the prediction and decision-making procedure so that users can feel confident and trust the system. Among various types of visual explanation formats, augmented reality-based explanations and product displays also enhance end users' trust in a system (Rodriguez-Sampaio et al., 2022). Too much information could create cognitive overload and decrease users' understanding and trust (Cramer et al., 2008; Schmidt et al., 2020; Hudon et al., 2021). The explanation should be stakeholder-oriented, such as by designing an interactive user interface to explain to non-technical stakeholders (Andres et al., 2020; Liao et al., 2020). Therefore, increased user interaction by providing adequate instructions and allowing the user to take initiatives would increase reliability and trustworthiness (Dodge et al., 2019; Oh et al., 2018; Putnam and Conati, 2019; Schrills and Franke, 2020).

If a system can simulate human-like expressions using lip sync and body language, it can increase trust (Weitz et al., 2021). For example, virtual assistants' voices, facial expressions, and gestures enhance users' trust. Therefore, a harmonic combination of a human-like facial expression along with an appropriate linguistic representation can have a significant impact on users' trust (Weitz et al., 2019; Gao et al., 2022). From the organizational point of view, employees' trust in any AI-based system is related to effectiveness, job efficiency, data protection, user understanding, and control. Weitz et al. (2019) also observed that though explanations should show the user relevant data along with the attributes, personal data need to be masked for privacy reasons (Wang and Moulden, 2021). Similarly, explanations that include comparisons among different attributes and previous and current recommendations can increase user (students, teachers, and educational researchers) trust in education-related XAI systems (Khosravi et al., 2022). Another study related to human resource management revealed that decreasing the knowledge gap between the user and the system can enhance trust (Chromik et al., 2021). Therefore, the authors also recommended reducing the knowledge gap by collaborating with users during the XAI

**Table 8**
Future research agenda.

| Themes | Issues/topics/research gaps | Challenges and research questions | Proposed approach and research paths |
| --- | --- | --- | --- |
| XAI Standardization | Lack of holistic guidelines for XAI development for researchers and practitioners.<br>XAI development process is opaque.<br>Research from a regulatory and compliance perspective is not available.<br>Communication methods and nature among the stakeholders is not defined. | Our review shows no empirical study that provides holistic guidelines or standards for developing an XAI System.<br>GDPR is newly introduced and one of the strictest guidelines. Hence, integrating it into XAI requires rigorous investigation.<br>RQ 1. How can XAI development guidelines be identified? | Alignment of the current software development cycle with XAI development.<br>Different stakeholders can be involved in the XAI software development lifecycle to determine the best practices/guidelines.<br>Identify the stakeholders.<br>Conduct qualitative and quantitative research to identify the stakeholder requirements.<br>Identify different types of stakeholder engagement with the XAI development lifecycle through qualitative and quantitative research.<br>Research across multiple domains can also help portray domain-specific guidelines as well as generic guidelines for XAI development. |
| | | RQ 2. How can we incorporate GDPR as a design requirement of XAI development? | Understand the applicable GDPR articles.<br>Codesign with industry practitioners, end users, and legal experts to investigate the GDPR requirements.<br>Conduct data protection impact assessments to evaluate the GDPR compliance trends.<br>Conduct design science research to identify common guidelines for GDPR compliance. |
| | | RQ 3. How do we integrate the proposed "Ethics guidelines for Trustworthy Artificial Intelligence" by "High-level Expert Group on Artificial Intelligence" from the European Commission?<br>RQ 4. How can the XAI stakeholders communicate with others for collaborative development? | Investigate and identify the feasibility of integrating ethical guidelines.<br>Co-design with practitioners and experts to outline the suitable ethical guideline requirements.<br>Stakeholders should be identified.<br>Codesign with the stakeholders to establish a collaborative development environment.<br>Use iterative evaluations of different types of communication techniques to find the suitable one. |
| XAI Visualization | Very few theory-guided studies have been conducted.<br><br>The measurement of information (or explanation) quality dimensions related to XAI are not discussed.<br><br>The dimensions of information (or explanation) quality are not discussed in detail for the low literacy group of people. | Information (or explanation) quality dimensions are not properly aligned with any current XAI literature.<br>User perception measurements of explanation quality have not been performed.<br>People with relatively low literacy might perceive the explanations differently. | Investigate the information quality dimensions in prior literature and align them with XAI.<br>Conceptualize XAI systems' explanation quality dimensions.<br>Develop or adapt explanation quality dimension measurement scales.<br>Use a theory-guided approach to explain how the explanation quality dimensions affect various outcomes. |
| | | RQ 5. How do we measure the explanation quality dimensions presented by XAI?<br>RQ 6. How does XAI representation differ in the case of relatively low literate people? | Evaluate and understand the need for differences in XAI representation for the low literacy group.<br>Conceptualize different XAI scenarios and present explanations in various formats.<br>Evaluate and measure various representations to find suitable representation techniques for XAI. |
| XAI Effects | Lack of measurement approach for trust, transparency, understandability, and usability for XAI systems.<br><br>Longitudinal study has not been performed.<br><br>Very few studies measure XAI's impact on domain experts, system developers, practitioners, and researchers. | Measuring the impact of different XAI representation formats.<br>XAI effect on low literacy group is not discussed.<br>The effect of AI explanation can have long-term effects; however, they are not outlined. | Design and validate measurement scales. |
| | | RQ 7. How do we measure user trust, transparency, understandability, and usability?<br>RQ 8. How do explanation quality dimensions affect trust, transparency, understandability, and usability? | Investigate the effect of explanation quality dimensions on trust, transparency, understandability, and usability through survey or |

**Table 8** (*continued*)

| Themes | Issues/topics/research gaps | Challenges and research questions | Proposed approach and research paths |
|---|---|---|---|
| | | RQ 9. How do we measure the XAI impact on different stakeholders? | experiment-based research. Behavioral theories can be used to propose suitable research models. Identify different stakeholders of the XAI ecosystem. Codesign with different stakeholders of XAI to understand the effect of explanations. Design measurement scale and collect user responses from different stakeholders. |
| | | RQ 10. What is the effect of XAI on low literate people? | Conceptualize case study suitable for people with low literacy. |
| | | RQ 11. What is the longitudinal effect of XAI on end users? | Understand how the effect of XAI changes over time among different stakeholders. Investigate user response over time and in different domains to measure the longitudinal effect. |

development lifecycle (Chromik et al., 2021; Hong et al., 2020; Park et al., 2021).

*5.1.2.2. Transparency.* Transparency denotes the concept of revealing the opaque procedure of decision making, allowing the whole procedure to be scrutinized by non-technical/average users if needed (Birkinshaw, 2006; Black, 1997). Therefore, making a process transparent can help to determine the features responsible for decision making, regulating, and controlling the whole process. To promote transparency, different attributes, such as age, gender, income, profession, and other related specifics, can be included in the explanation (Janssen et al., 2020). For autonomous cars, there is a significant tradeoff between explainability and system complexity. If the system needs to be more transparent (explainable), the design becomes complex, and the whole process becomes time-consuming as well (van der Waa et al., 2020).

For movie recommendation systems, content-based collaborative filtering can be adopted to increase transparency. Therefore, item-based recommendations and user-centric recommendations should be distinguishable so that the user is informed about the system and can easily connect the dots between their expectation and the system-provided recommendations (Conati et al., 2021; Ngo et al., 2020; Li et al., 2021; Liu et al., 2021a, 2021b; Dhanorkar et al., 2021). Along with the explanation, contextual information and reference data are also helpful in promoting the transparency of the systems (Liao et al., 2020). A user's involvement in the system development process by facilitating reliable communication between the system development team and the user increases transparency (Cai et al., 2019; Eiband et al., 2018). Therefore, transparency in the decision-making process and user involvement in system development can positively impact system acceptance (Conati et al., 2021; Cramer et al., 2008).

*5.1.2.3. Understandability.* Experimental research shows that a user's prior knowledge about the system's interactions results in better understandability and trust in the system (Branley-Bell et al., 2020; Cheng et al., 2019; Eslami et al., 2018; Lim et al., 2009; Bove et al., 2021). Moreover, presenting every interaction within the system in a sequential manner helps users understand the working procedure of the system (Broekens et al., 2010). Ehsan et al. (2021) found that social transparency is important in increasing an AI-based system's understandability; however, without background information and proper contextual information, the prediction accuracy (confidence measurement) is nothing but a number (Ehsan et al., 2021; Bove et al., 2021). Explanations with logical reasoning and counterfactual information improve the understandability of the system (Górski and Ramakrishna, 2021). In the case of expert-level end users, counterfactual explanations help to understand the generated explanations, decision making, and factors relevant to the algorithms (Evans et al., 2022). Case-based explanations can increase the understandability of decision making in criminal justice related use cases (Liu et al., 2021a, 2021b). The study by Liu et al. (2021a, 2021b) also showed that if users complete some training before using a system, this can increase the interactive nature of and the familiarity with the explanations (Liu et al., 2021a, 2021b).

Binns et al. (2018) found that a complete view of an explanation with strategic details of each process increases the user's understandability of a system. Here, the complete view denotes that the explanation should be more comprehensive and should include the scope of each event that takes place within the system's boundaries (Binns et al., 2018; Ehsan et al., 2019). Apart from textual explanations, visual explanations of the input data of the machine-learning model promote a higher level of visibility, understandability, observability, and trust in a system (Lim and Dey, 2009; Schrills and Franke, 2020; Daudt et al., 2021). Explanations involving augmented reality can also impact average users' understandability of the system (Rodriguez-Sampaio et al., 2022). In addition to an automated explanation system, an on-demand feedback retrieval system enhances the understandability of autonomous vehicle decision making (Schneider et al., 2021). Moreover, for wearable systems, auditory feedback of explanations increases a system's understandability (Danry et al., 2020).

In the case of non-technical stakeholders, the information should be clear, concise, and comprehensive so that there is no unnecessary information that might create a cognitive overload (Hudon et al., 2021). Users of AI-based hiring systems require numerical data of the assessment along with the explanation to increase understandability. The attributes should be properly labelled and explained, and the decision should be properly reasoned to increase user understandability (Li et al., 2021). Furthermore, the explanation provision can be on-demand to avoid the monotonous and time-consuming nature of a system (Chazette and Schneider, 2020). Another way to increase understandability is to use fact sheets and mental models for a variety of stakeholders involved in the AI development process (Chromik et al., 2021; Hind et al., 2020). The fact sheet contains all the attributes of data, the prediction mechanism, the working principle, the inherent structure of the model, the training data for machine-learning models, testing protocols, and testing models (Hind et al., 2020). In addition, the developers of XAI must understand the user's mental model before developing the system. Mental model is crucial for any interactive system design since it is based on users' beliefs and perceptions about the external world. Therefore, the developer team must collaborate with end users, domain experts, and other necessary stakeholders to establish dedicated communication (Chromik et al., 2021).

*5.1.2.4. Usability.* XAI systems can have a positive impact on a system's usability (Oh et al., 2018). According to Chazette and Schneider (2020), for navigation systems, users would like to feel in control of the system because it provides the user with the choice of accepting or rejecting a decision. Furthermore, feedback modalities/features in autonomous vehicles can significantly increase user experiences by making the system more usable and understandable (Chazette and Schneider, 2020).

For the finance and human resource domain, following a particular explanation style is vital because presenting various explanations using a specific explanation style could help the user understand the role of various features and the reasoning behind the prediction, which can improve usability. Similarly, Szymanski et al. (2021) found that in the case of a news article recommendation system, explanations help users assess their own article writing skills and at the same time learn to improve their articles. Furthermore, to increase usability, accessible and interactive interfaces should be designed and developed for non-technical stakeholders (Andres et al., 2020; Brennen, 2020). Involving the stakeholders in the development lifecycle may also increase a system's usability (Chromik et al., 2021).

*5.1.2.5. Fairness.* The fairness of an intelligent system is dependent on various attributes and values as well as validity. Local explanation, which refers to an explanation of each prediction, enhances system fairness perceptions for the user. Case-based explanations have less impact on fairness criteria, but a global explanation can compensate for this and can enhance user trust (Dodge et al., 2019); however, for a criminal justice use case-based explanation, evidence-based-reasoning, legal rule sentences, and citation sentences have impacts on users' fairness perceptions (Liu et al., 2021a, 2021b; Górski and Ramakrishna, 2021). Social media related health applications and services require explanations with all types of details, logical reasoning, demographic information, and supplementary information to increase fairness perceptions among users (Liu et al., 2021a, 2021b).

For the finance and human resource domains, explanation style is vital to fairness perceptions. As mentioned, the different explanations presented in similar explanation styles could help the user understand the role of various features and the reasoning behind a prediction (Binns et al., 2018). Therefore, the user's ability to differentiate among various reasons will increase, resulting in enhanced fairness perceptions (Binns et al., 2018; Janssen et al., 2020). Furthermore, fairness is perceived more favorably by users when the input influence explanation presented is understood (Binns et al., 2018; Mucha et al., 2021).

*5.1.3. Explanation presentation time*

Our critical observation of the selected literature shows that explanations are provided to users in two ways. In most cases, the explanation is shown to the user automatically while visualizing the decision itself. In this case, the user wants minimal and adequate information to be provided to avoid cognitive overload (Ehsan et al., 2019; Schmidt et al., 2020; Hudon et al., 2021; Dhanorkar et al., 2021). Therefore, to have more control over an explainable AI system, users also prefer on-demand supplementary and contextual information sharing. Hence, "when the AI should be explainable" revolves around these primary concepts. For different domains, the concept is presented in various ways because the nature of the interaction is not the same across all domains. Some of these scenarios are described in detail.

For medical personnel, both textual and visual explanations and related hints are displayed automatically to doctors after decision making (Branley-Bell et al., 2020; Eiband et al., 2018, 2019; Xie et al., 2019). In addition, supplementary information should be available upon user request for better diagnosis, understandability, and trust (Wang et al., 2019; Xie et al., 2019). The supplementary information can be a combination of a reference to the previous diagnosis or any historical data that might help make an accurate decision (Branley-Bell et al., 2020; Bussone et al., 2015; Daudt et al., 2021; Lee and Rich, 2021). Similarly, various media and entertainment systems require textual and visual explanations immediately with the prediction result (Kouki et al., 2019; Ngo et al., 2020). Music recommendation systems, art recommendation systems, arcade gaming, tour guides, cooking agents, and movie recommendation systems require automatic rational generation along with personalized recommendations (Cramer et al., 2008; Ehsan et al., 2019; Lim and Dey, 2009; Ngo et al., 2020; Schmidt et al., 2020). Another requirement for automatic explanations is to reveal the filtering technique used as well as which data have been considered. For AI-based drawing tools, the user needs information on-demand rather than automatic explanations because users need to lead the task rather than receive suggestions from the system (Oh et al., 2018). One demand explanation is also required for intelligent cooking agents. The users like to lead the task and later like to receive explanations from the system (Broekens et al., 2010).

In the case of the financial domain, users are automatically presented with an explanation regarding decision making. Though the decision making is automatic, the system should show the explanation in different styles to make it more understandable (Binns et al., 2018; Cirqueira et al., 2020). Similar to healthcare decision making, contextual information is also needed upon user request (Chromik et al., 2021; Rodriguez-Sampaio et al., 2022). Autonomous car users require automatic and prompt textual and visual explanations along with the prediction result for quick decision making (Chazette and Schneider, 2020; Schneider et al., 2021). Contextual information about different scenarios should be available upon user request. Moreover, the literature analysis revealed that users of human resource management, e-commerce, and other recommendation systems (Broekens et al., 2010; Conati et al., 2021; Ehsan et al., 2021; Park et al., 2021; Zimmermann et al., 2022) require both on-demand and automatic explanations. Therefore, it is clear that in most cases, users receive explanations automatically along with the prediction result; however, supplementary information is necessary and should be available on-demand in most cases. The supplementary information can be textual, visual, or hybrid because there is no clear infrmation of this requirement in the literature. In addition, job recruiters or recruitment agencies sometimes need to backtrack the decision they made to get help in the next recruitment. Backtracking helps to understand the decision-making process of the system. Therefore, human resource managers and recruiters require mostly on-demand explanations (Li et al., 2021; Daudt et al., 2021).

## 5.2. Research domains

We have identified 10 domains in which XAI has been used: healthcare, media and entertainment, education, transportation, finance, e-commerce, human resource management, digital assistant, e-governance, and social networking. We comprehensively discuss the use of XAI in these domains in more detail.

### 5.2.1. Healthcare

Healthcare is one of the most explored research domains in XAI. This domain includes research on clinical decision making, disease diagnosis, and health-related recommendation systems (Branley-Bell et al., 2020; Bussone et al., 2015; Cai et al., 2019; Wang et al., 2019; Rodriguez-Sampaio et al., 2022). The users of AI-based systems in healthcare are primarily doctors with very little technical knowledge. Moreover, they tend to have their own opinion regarding disease detection and clinical decision making (Branley-Bell et al., 2020; Bussone et al., 2015; Lee and Rich, 2021). Therefore, the explanation required by doctors should include sufficient graphical and textual data along with appropriate contextual references (Branley-Bell et al., 2020; Daudt et al., 2021; Xie et al., 2019). In addition, healthcare-based applications, such as fitness apps and nutrition recommendations, should have a communicative and interactive user interface (Eiband et al., 2018, 2019).

### 5.2.2. Media and entertainment

This research domain consists of music recommendation systems, movie recommendation systems, art recommendation systems for museums and websites, news article recommendation systems, and arcade gaming systems (Kouki et al., 2019; Ngo et al., 2020; Oh et al., 2018; Szymanski et al., 2021). Users of both music and movie recommendations prefer personalized recommendations presented in various explanation styles (Ngo et al., 2020; Schmidt et al., 2020). They require explanations containing the details of the personalized recommendations that include the basic working principle of the system and details regarding the personal information used (Ngo et al., 2020; Schmidt et al., 2020). Users also express their concern regarding the amount of information being presented because too much information can create cognitive overload (Ehsan et al., 2019; Schmidt et al., 2020; Hudon et al., 2021). Therefore, the system should not overwhelm the user with unnecessary and ambiguous information and should provide both textual and visual explanations. In the case of gaming, the user interface should provide prompt hints, and the user interface should be communicative and user-friendly (Cramer et al., 2008; Ehsan et al., 2019; Schmidt et al., 2020).

### 5.2.3. Education

The education domain includes intelligent tutoring systems, university admission decision making, and grade estimation systems (Cheng et al., 2019; Conati et al., 2021; Mucha et al., 2021; Putnam and Conati, 2019). Investigations on intelligent tutoring systems have revealed that explanations improve the usability of the system (Conati et al., 2021; Putnam and Conati, 2019). In addition, the explanations should provide information about the system's behavior and working procedure as well as the logic behind certain decision-making tasks, such as admission decision making. For admission decision making, users often argue it should be humans who should make the decision, not a machine that simply runs on algorithms (Cheng et al., 2019; Mucha et al., 2021; Khosravi et al., 2022).

### 5.2.4. Transportation

The transportation domain includes navigation systems, decision support systems for autonomous cars, and flight rerouting systems for the aviation industry (Binns et al., 2018; Chazette and Schneider, 2020; Schneider et al., 2021; van der Waa et al., 2020). For domain experts, case-based explanations are preferable for autonomous car decision support systems (Schneider et al., 2021). Case-based explanations refer to explaining a certain decision in relation to use cases (Schneider et al., 2021). Moreover, the explanations (hints) provided to the user can be visual, textual, light indicators, or a hybrid mode (Schneider et al., 2021; van der Waa et al., 2020). For navigation systems, the user requirements are slightly different because the users require on-demand explanations as well as proper reasoning behind any decision being made. A similar scenario is observed in the case of flight re-routing systems. In both cases, the users wanted to control the flow of suggestions (explanations) provided by the system (Binns et al., 2018).

### 5.2.5. Finance

Financial use cases of XAI research include insurance, financial fraud detection, and loan applications (Binns et al., 2018; Chromik et al., 2021; Cirqueira et al., 2020). For banking activities, such as insurance claims and loan approvals, explanations regarding specific decision making should be made available to users (Chromik et al., 2021). The operator should be able to see the loan requestor's information, credit history, and other demographic information. Binns et al. (2018) and Cirqueira et al. (2020) also argued that when designing an explainable system, the developers must understand and connect with the user's mental model. An effective XAI system should be able to detect the incorrect mental model and calibrate it accordingly (Binns et al., 2018; Cirqueira et al., 2020).

### 5.2.6. E-commerce

The authors have investigated the social transparency and design framework that contributes to trust in the decision making of AI-based systems (Ehsan et al., 2021; Eslami et al., 2018) used in e-commerce. A better understanding of artificial intelligence-based systems is required for the promotion of social transparency. Although the effect of transparency of XAI in the long term has not been investigated, it can be utilized as a useful marketing tool (Eslami et al., 2018). Furthermore, because online advertising uses personal data for analytics, a less transparent algorithm may increase privacy issues (Ehsan et al., 2021). Online shopping experiences are better if there is an explainable AI-based system using augmented reality and text. The users have more trust and therefore a better online shopping experience (Zimmermann et al., 2022; Bove et al., 2021).

### 5.2.7. Human resource management

Employees' attitudes towards accepting artificial intelligence-based decisions in human resource management are influenced by a variety of factors (Binns et al., 2018; Park et al., 2021; Bankins et al., 2022). Employees often believe that the decision making can be biased, manipulative, and an invasion of privacy. Therefore, a psychological burden in accepting AI-based predictions could result. Reducing the knowledge gap by increasing transparency and interpretability can help in understanding decision making (Park et al., 2021). Moreover, collaborating with human users during the design stage can increase the chances of system adoption and can enhance the positive attitude toward the system (Binns et al., 2018; Park et al., 2021). The algorithmic hiring process is becoming increasingly popular. The recruiter's requirements for explainability include providing appropriate reasons behind the decision making, explaining the assessment scores of the candidate, and showing similar recruitments in the organization. A previous recruitment of similar kind can help the recruiters detect any possible bias in the decision-making process. Moreover, if multiple recruiters are using the system and are changing shifts, it is a good idea to provide a summary of previous work each time they log in to the system (Li et al., 2021).

### 5.2.8. Digital assistants

Previous studies have shown that the more human-like and interactive the system is, the more user trust increases for virtual assistants (Weitz et al., 2019, 2021). Facial expressions, voice, gestures, and verbal comments, especially those related to phonemes, are supportive and appealing to users. Moreover, end users require linguistic explanations from an XAI system. Hence, an interactive agent with a harmonic combination of explainable AI methods and an appropriate linguistics representation can make a system trustworthy and more user-centered (Gao et al., 2022; Weitz et al., 2021).

*5.2.9. E-governance*

Empirical analyses have been performed on a criminal justice use case to investigate people's perceptions of the fairness of machine-learning algorithms and to what extent these algorithms need explanations (Dodge et al., 2019; Janssen et al., 2020). To increase understandability, credibility, and trust, the system should explain the algorithm's working procedure, the attributes that contribute to decision making, and the availability of contextual data (Dodge et al., 2019). Similarly, investigations of immigration services use cases reveal that though algorithms can help in decision making, it is not necessary to make all decisions using algorithms (Janssen et al., 2020). One study also revealed that the white box approach (explainable AI approach) can lead to better decision making (Janssen et al., 2020). Therefore, e-governance requires human intervention for critical decision making.

*5.2.10. Social networking*

Research on the social networking domain has revealed that the participants require both "why" and "why not" explanations for specific system behaviors (Lim and Dey, 2009; Yin et al., 2019; Liu et al., 2021a, 2021b). Therefore, developers can provide user log information, mental model related information, and contextual information on-demand. Moreover, an effective explainable AI system requires human user intervention in the design process through a dedicated communication medium (Yin et al., 2019).

Apart from the application domains of XAI, several studies have discussed the realm of XAI development. Studies related to XAI have been conducted to develop practical guidelines for designers, developers, domain experts, and other related stakeholders (Hind et al., 2020; Hong et al., 2020; Liao et al., 2020; Wang and Moulden, 2021). Hind et al. (2020) designed a question bank as a standard guideline for collecting user requirements for user-centered AI. The guidelines provided in this study can be a vital component in designing a trustworthy, understandable, interactive, and user-centric XAI system (Hind et al., 2020). Developers should also explore the problem space and conceptualize primary and alternative strategies (Hong et al., 2020; Liao et al., 2020). In addition, XAI development requires the active participation of domain experts, product managers, data scientists, auditors, and end users (Wang and Moulden, 2021).

## 6. Critical analysis of future research agendas

This section focuses on questioning and problematizing future research directions (Alvesson and Sandberg, 2011, 2020). In contrast to the previous section's discussion of the XAI research trend, this section extensively focuses on establishing a critical standpoint of future research directions by analyzing "what" is the current knowledge and "how" it can be improved (Alvesson and Sandberg, 2011, 2020). Therefore, we have reconsidered the current understanding related to XAI's methodological, conceptual, and development issues and investigated the unexplored areas. We have divided the whole observation into three primary thematic categories. The first one considers the standardization practice, the second focuses on representing XAI, and the last considers the overall effect of XAI on humans. Furthermore, rather than simply pointing out the gap that exists in the research findings, we have tried to articulate emerging research questions deduced from the unexplored research areas. We then constructed them in terms of their potential significance to identify specific and feasible research paths. Table 8 provides an overview of the future research directions based on current knowledge.

### 6.1. Theme 1: XAI standardization

Our analysis reveals that XAI has been used in various domains; however, there is a lack of studies that inform XAI standardization. One of the articles provides the guidelines for UI design for XAI, which both the designers and developers can use if needed (Eiband et al., 2018). Another article proposes a question bank that might be useful for requirement elicitation for explainable AI (Liao et al., 2020); however, these two articles do not offer comprehensive guidelines or standards for developing an explainable AI system. Therefore, the following research questions can be addressed for the XAI standardization theme.

*6.1.1. RQ 1. How can XAI development guidelines be developed?*

Extensive research on XAI design and development can facilitate determining standard guidelines and best practices for XAI development. Therefore, an important research direction would be identifying the best practices and guidelines for XAI development. Addressing this research question should include the involvement of all necessary stakeholders in the research. Furthermore, researchers across multiple domains can help create domain-specific guidelines for XAI development. Design science research can also be used to create XAI development guidelines (Hevner and Chatterjee, 2010).

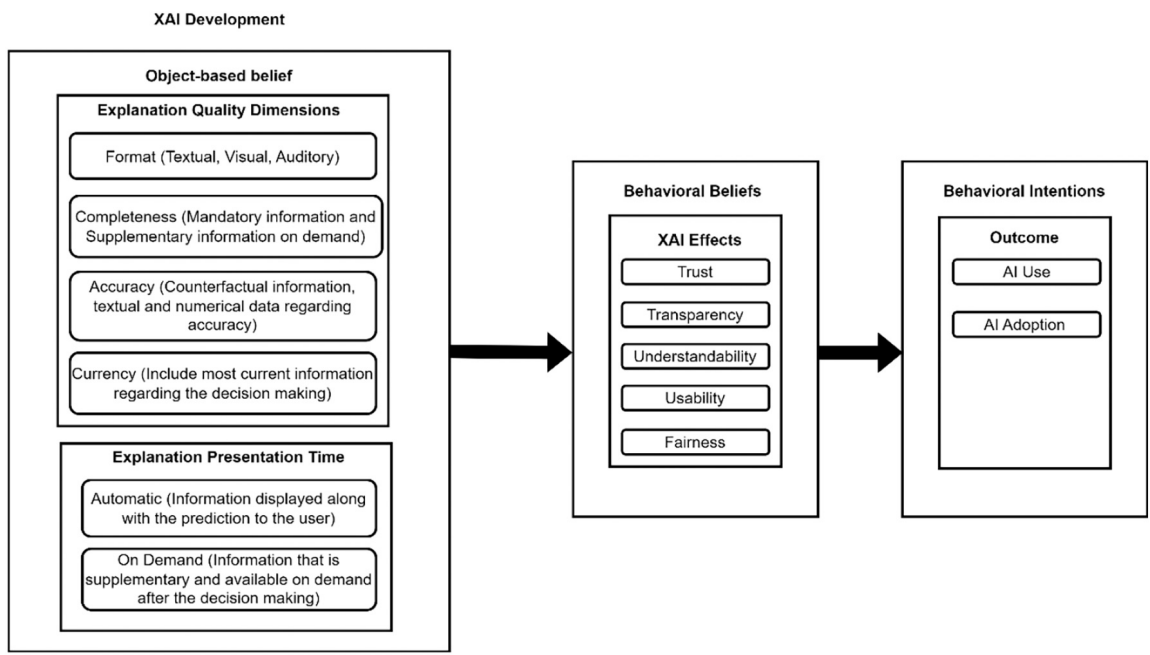

**Fig. 2.** Synthesized framework for XAI research from a user perspective.

*6.1.2. RQ 2. How can we incorporate regulatory and ethical aspects as design requirements of XAI development?*

Our observation reveals a lack of empirical studies from a regulatory and compliance perspective. Article 22 of the GDPR discusses "automated individual decision-making, including profiling," to safeguard the data subject's personal information from automatic processing (Malgieri, 2019; European Union, 2018). In addition, Articles 13–15 of the GDPR discuss the data subject's right to know the logic, that is, "meaningful information," regarding the processing of personal data. To be more precise, the data subject has the right to be informed about "meaningful information about the logic involved" if any decision related to the subject is "based solely on automatic, automated processing" (Malgieri, 2019).

To address this research question, researchers must identify the possible GDPR articles related to the XAI system. The requirements of GDPR compliance are mostly related to personal data collection, processing, retention strategy, and destruction. Therefore, co-design work that involves regulators, auditors, privacy offers, and other necessary stakeholders is a useful research direction. Another crucial step is to conduct a data protection impact assessment.

*6.1.3. RQ 3. How can the stakeholders communicate with the developer team for XAI development?*

We have observed from the review that communication among the developer team and other stakeholders are essential to XAI development; however, there are limited guidelines to initiate and conduct such communication (Meske et al., 2022). Therefore, to address this research question, the researchers can organize co-design workshops with the stakeholders of the XAI ecosystem so that the communication techniques can be identified and evaluated.

*6.2. XAI visualization*

*6.2.1. RQ 4. How do we measure the explanation quality dimensions of XAI?*

We have identified explanation quality dimensions in this paper. Previous studies did not empirically measure the explanation quality dimensions. In our review, we also observed that the explanation quality dimensions of AI systems unfold differently than the information quality dimensions. Therefore, researchers can search for the availability of existing measurement scales for format, completeness, accuracy, and currency. If such scales are available, researchers can adapt them to the XAI context. A major adaptation of these scales would be needed, and in fact, researchers may need to develop the scales from scratch by following the standard scale development procedure (Moore and Benbasat, 1991).

*6.2.2. RQ 5. How does explanation representation differ in the case of relatively low-literate people?*

The low-literacy group tends to have low AI literacy, which makes information representation more challenging. The selected articles used in this work revealed textual, visual, auditory, and hybrid modes of information representation. Different modes are used for different application domains; however, no article investigates neither how to represent the explanations to low-literate people nor how to measure their perceptions of the explanation. Therefore, addressing this research question can help present an explanation suitable for all users. The researchers should evaluate different XAI representations among low-literate people. It is vital to conceptualizing various XAI scenarios that can be presented to them.

*6.3. XAI effects*

*6.3.1. RQ 6. How do we measure the trust, transparency, understandability, and usability of XAI? How do explanation quality dimensions affect trust, transparency, understandability, and usability?*

Our observation in this review work revealed that a limited number of studies exist that measure the user perceptions of the transparency, understandability, and usability of an XAI system (Cramer et al., 2008; Daudt et al., 2021; Cheng et al., 2019). Thus, researchers should use existing measurement scales to measure these factors. The identified scales should be adapted to the XAI context. Theory-guided approaches can be used to construct models to investigate how explanation quality affects satisfaction, trust, transparency, understandability, and usability.

*6.3.2. RQ 7. How do we measure the XAI impact on different stakeholders?*

An AI ecosystem contains different stakeholders, such as designers, domain experts, developers, data scientists, UX engineers, and regulatory bodies (Meske et al., 2022; Laato et al., 2022). For example, domain experts can participate in the XAI development process to identify the feasibility of the explanations. Data scientists can assist the development process by designing more explainable machine-learning models. Similarly, other stakeholders can contribute to XAI development. To understand the impact on different stakeholders, a similar methodological approach can be adopted as we suggested in RQ6.

*6.3.3. RQ 8. What is the effect of XAI on low-literate people?*

We did not find studies that targeted low-literate people. To address this research question, first, researchers need to identify the low-literate group of people. Experiments can be designed in which AI decision-making and explanations can be presented to collect responses on explanation quality and other important factors. This type of research can also validate the developed scales in RQ4 and RQ6 among low-literate user groups.

*6.3.4. RQ 9. What is the longitudinal effect of XAI on various types of end users?*

Human-centered XAI can benefit from longitudinal studies because it will help researchers understand the changes in user perceptions overtime at the group level and individually. Most prior research studies on explainable AI are cross-sectional. Researchers can develop relevant research models and test them using longitudinal research design.

**7. Synthesized framework for XAI research from users' perspectives**

The findings from the current SLR enabled us to construct a comprehensive framework for XAI research from end users' perspectives (Fig. 2). Building on the work of Wixom and Todd (2005), our proposed comprehensive framework suggests that object-based beliefs, such as the explanation quality dimensions (format, completeness, accuracy, and currency) as well as when to explain (automatic and on-demand), impact a number of behavioral beliefs, including trust, transparency, understandability, usability, and fairness. In turn, these behavioral beliefs impact behavioral intention (AI adoption, AI use).

According to Wixom and Todd (2005), object-based beliefs are the characteristics of technology, whereas behavioral beliefs are the anticipated consequences of technology use. Wixom and Todd (2005) suggested that the impacts of object-based beliefs on behavioral beliefs are mediated through the object-based attitude (Eagly and Chaiken, 1993; Fazio and Olson, 2003); however, in a recent empirical study (Islam et al., 2020), it was shown that object-based beliefs can have direct impacts on behavioral beliefs. Therefore, we have proposed direct relationships between object-based beliefs and behavioral beliefs in our framework. Fig. 2 shows the graphical representation of the framework.

## 8. Implications

*8.1. Theoretical implications*

Our SLR findings have five major theoretical contributions. First, from a broad perspective, our study is one of the few studies investigating AI end users' explanation needs. Therefore, our paper contributes to the previously conducted literature reviews (Wells and Bednarz, 2021; Anjomshoae et al., 2019; Gerlings et al., 2021a, 2021b; Laato et al., 2022), particularly by identifying the end users' explanation needs and the impacts.

Second, we adopted Wixom and Todd's (2005) conceptualization of information quality dimensions to conceptualize the explanation quality dimensions of AI systems. Our findings show that explanation quality dimensions are format, completeness, accuracy, and currency. We have also observed that when to explain (automatic and on-demand) is another important factor of XAI. With our findings, we contribute to research conducted to design and govern responsible AI systems (Wearn et al., 2019; Maas, 2018; Peters et al., 2020; Rakova et al., 2021).

Third, we have described the five effects of XAI systems: trust, transparency, understandability, usability, and fairness. Our SLR findings position these factors as the most important effects of XAI. While these factors are described by Laato et al. (2022), our SLR links them with XAI representation dimensions.

Fourth, we have identified three major themes of future research: XAI standardization, XAI visualization, and XAI effects. We have proposed nine possible research questions that future IS researchers can investigate. We have also outlined the possible ways researchers can address these research questions.

Finally, we have proposed a comprehensive framework by connecting explanation-related factors and XAI effects. We further propose that the XAI effects can ultimately influence behaviors, such as AI adoption and use. This framework has implications for researchers. For example, many interesting research models can be developed and tested based on this framework. While the framework is developed using Wixom and Todd's (2005) work, which describes relationships among object-based beliefs, behavioral beliefs, and behavior, hypotheses can be developed from additional theories, such as the IS success model (DeLone and McLean, 1992, 2002), technology acceptance models (Davis, 1989; Davis et al., 1989; Chuttur, 2009), the theory of reasoned actions (Ajzen and Fishbein, 1973; Ajzen and Fishbein, 1980; Fishbein, 1967; Fishbein and Ajzen, 1977; Hale et al., 2002), and the theory of planned behavior (Ajzen, 1985, 1991).

*8.2. Practical implications*

From a practical standpoint, this SLR can serve as a guideline for designing human-centric AI and measuring its consequences. Because AI is becoming more prevalent in all aspects of life, the findings of this study may drive researchers and enthusiasts to design digital services that are morally sustainable. For example, designers can ensure that their systems provide explanations related to the identified dimensions of explanation quality. Their design should also contain possibilities for both automatic and on-demand explanations. Our findings also outline a need to design XAI systems in various domains, not just for mission-critical systems. Since AI is now being used more than ever in various industrial and corporate decision-making, the findings of this SLR can help understand the employees' behavioral intention to use those systems. As discussed in the literature, various state-of-the-art recruitment systems use data-driven decision-making. In cases like this, the explanation dimensions can help to understand the details of the decision-making process. Hence, the synthesized framework of this SLR can be adopted in various industries and corporate organizations to understand the likelihood of system adoption and use. Therefore, we suggest that system designers consider this need when they design AI-based systems. This also has implications for AI education. We suggest including topics such as explainable AI, responsible AI, and AI governance, among others, as important topics to train AI developers in addition to technical topics.

## 9. Conclusion

Recently, AI has gained significant momentum, which, if correctly managed, does have the potential to revolutionize various sectors; however, the AI community must overcome the challenge of explainability, an intrinsic hurdle that was not a part of AI-based ecosystems before. This work has comprehensively discussed XAI from the end user's perspective. We have identified the dimensions of explanation quality from existing empirical studies, and we found that the effects of XAI on end users can motivate users to adopt and use AI-based systems. Furthermore, by investigating the selected studies, we have identified crucial future research avenues. Possible directions to address these avenues and a comprehensive framework have also been identified and developed, respectively. Though the widespread application of XAI is yet to be implemented, based on our review, the growing need for XAI is vividly clear. The explanation quality dimensions of XAI outlined in this work are vital to XAI system development because the dimensions can have impacts on trust, understandability, fairness, and transparency.

Our study has three limitations. First, we have considered only the empirical studies on XAI for this review work. Future studies can also consider theoretical papers on XAI.

Second, we used Scopus and Web of Science for the database search. Hence, we might have missed important studies for our work. This limitation can be addressed in the future by conducting searches of other databases.

Third, we have used Wixom and Todd's (2005) information quality dimensions for conceptualizing the explanation quality of AI systems. There are other information quality dimensions proposed by other researchers (Wang and Strong, 1996). Therefore, future studies can use these dimensions to identify additional explanation quality dimensions for AI systems.

**CRediT authorship contribution statement**

AKM Bahalul Haque: Conceptualization, Methodology, Conducting Primary Search, Data Collection, Writing Original Draft, Analyzing and Addressing the Reviewer's Comments

A.K.M. Najmul Islam: Conceptualization, Reviewing Draft, Editing, Reviewing the Search Result, Critically Analyzing Reviewers' Comments, Supervision

Patrick Mikalef: Conceptualization, Reviewing Draft, Reviewing the Search Result and Data, Critically Analyzing Reviewers' Comments, Supervision

**Declaration of competing interest**

None.

**Acknowledgements**

This work was supported by the Slovenian Research Agency (research core funding No. P5-0410).

**Appendix A. Supplementary data**

Supplementary data to this article can be found online at https://doi.org/10.1016/j.techfore.2022.122120.

**AKM Bahalul Haque** is a Junior Researcher at the Department of Software Engineering at LUT University. Earlier, he was a lecturer at the Department of Electrical and Computer Engineering, North South University. His works have been accepted and published in international conferences and peer-reviewed journals, including IEEE Access, Expert Systems, Cybernetics and Systems, various International conference proceedings, Tylor and Francis Books, and Springer Book. His research interests include Explainable AI, blockchain, data privacy and protection, and human-computer interaction.

**A.K.M. Najmul Islam** received the Ph.D. degree in information systems from the University of Turku, Finland, and the M.Sc. (Eng.) degree from the Tampere University of Technology, Finland. He is currently an Adjunct Professor at Tampere University, Finland. He is also an Associate Professor at LUT University, Finland. His research has been published in top outlets, such as European Journal of Information Systems, Information Systems Journal, Journal of Strategic Information Systems, Technological Forecasting and Social Change, Computers in Human Behavior, Internet Research, Computers & Education, Journal of Medical Internet Research, Information Technology & People, Telematics & Informatics, Journal of Retailing and Consumer Research, Communications of the AIS, Journal of Information Systems Education, AIS Transaction on Human-Computer Interaction, and Behaviour & Information Technology.

**Patrick Mikalef** is a Professor in Data Science and Information Systems at the Department of Computer Science. He has been a Marie Skłodowska-Curie post-doctoral research fellow working on "Competitive Advantage for the Datadriven Enterprise" (CADENT). He received his B.Sc. in Informatics from the Ionian University, his M.Sc. in Business Informatics for Utrecht University, and his Ph.D. in IT Strategy from the Ionian University. His research interests focus on the strategic use of data science and information systems in turbulent environments. He has published work in international conferences and peer reviewed journals, including the European Journal of Information Systems, British Journal of Management, Information and Management, and the European Journal of Operational Research.